\documentclass[letterpaper, 10 pt, conference]{ieeeconf}  

\IEEEoverridecommandlockouts                              

\usepackage{graphicx} 
\usepackage{hyperref} 

\title{\LARGE \bf
MVLidarNet: Real-Time Multi-Class Scene Understanding \\ for Autonomous Driving Using Multiple Views
}
\author{Ke Chen$^*$, Ryan Oldja$^*$, Nikolai Smolyanskiy$^*$, \thanks{$^*$Equal contribution.} Stan Birchfield\\Alexander Popov, David Wehr, Ibrahim Eden, Joachim Pehserl \\ NVIDIA}

\begin{document}

\maketitle
\thispagestyle{empty}
\pagestyle{empty}


\begin{abstract}
Autonomous driving requires the inference of \emph{actionable information} such as detecting and classifying objects, and determining the drivable space.
To this end, we present Multi-View LidarNet (MVLidarNet), a two-stage deep neural network for multi-class object detection and drivable space segmentation using multiple views of a single LiDAR point cloud.
The first stage processes the point cloud projected onto a \emph{perspective view} in order to semantically segment the scene.
The second stage then processes the point cloud (along with semantic labels from the first stage) projected onto a \emph{bird's eye view}, to detect and classify objects.
Both stages use an encoder-decoder architecture.
We show that our multi-view, multi-stage, multi-class approach is able to detect and classify objects while simultaneously determining the drivable space using a single LiDAR scan as input, in challenging scenes with more than one hundred vehicles and pedestrians at a time.  
The system operates efficiently at 150~fps on an embedded GPU designed for a self-driving car, including a postprocessing step to maintain identities over time.
We show results on both KITTI and a much larger internal dataset, thus demonstrating the method's ability to scale by an order of magnitude.\footnote{Video is at \url{https://youtu.be/2ck5_sToayc}}
\end{abstract}

\section{INTRODUCTION}

Autonomous driving requires perception of \emph{actionable information}, \emph{i.e.}, data that can be directly consumed by the subsystem that controls the vehicle.
Actionable information includes data such as the locations of the lanes, the curvature of the road, the color of the stop light, the presence of construction cones, whether the stop sign is facing the vehicle, the distance to the car in front, whether the pedestrian is crossing the road, and so forth.
Such information is immediately useful in determining whether the vehicle should turn, accelerate, or brake.

Of these various types of actionable information, perhaps the most attention has been paid to the detection of nearby cars, cyclists, and pedestrians.
To solve this problem, researchers have proposed methods using either RGB images, LiDAR data, or a fusion of the two modalities.  
While RGB object detection itself is relatively mature, lifting such results from image space to world space often yields significant geometric inaccuracies.
LiDAR data solves this problem but introduces another: directly processing a 3D point cloud is not straightforward.
One possibility is to process the point cloud as a 3D voxel grid \cite{luo2018cvpr:fastnfurious}, but this is computationally expensive and introduces quantization errors; another is to project the point cloud to a bird's eye view (BEV) as a height map or multi-channel representation \cite{yang2018cvpr:pixor}, but this loses potentially valuable information, especially for pedestrians due to their small size.


In this paper we propose to overcome these limitations by projecting a LiDAR input into both perspective (ego-centric) and top-down (bird's eye) views.
This provides the best of both worlds, since the former allows us to leverage shape information that is so crucial for detecting pedestrians, and the latter allows us to aggregate information in a format that is useful for autonomous driving.
To facilitate the detection of pedestrians, we leverage semantic segmentation of the LiDAR points, which has only been recently been made possible with the introduction of the SemanticKITTI~\cite{behley2019iccv:semantickitti} dataset that contains pointwise ground truth segmantic segmentation. 

Our approach leverages a two-stage multi-view network that performs semantic segmentation on a perspective-view projection of the point cloud, followed by object detection and classification on a bird's-eye projection.
Unlike previous approaches, we focus on \emph{multi-class} detection, in which the same network is used to detect multiple object classes simultaneously (that is, without training separate network weights for each class).
As we show in the experimental results, this simple approach, leveraging two encoder-decoder stages, is able to achieve competitive results on multi-class KITTI object detection, while simultaneously determining the drivable space.
The simple design yields efficient processing, enabling the system to process LiDAR point clouds at 150~fps with competitive accuracy, while maintaining identities over time via postprocessing.

%

The paper contains the following contributions:
\begin{itemize}
	\item We present a novel \emph{multi-view} \emph{multi-class} system to detect vehicles and pedestrians while simultaneously computing drivable space, all with a single network processing a LiDAR input via two different projected views.
	\item Due to the simplicity of its design, our system operates faster than previous approaches, at 150~fps on an embedded GPU. 
	\item Results of the system are shown on extremely challenging data with more than one hundred vehicles and pedestrians per frame, thus advancing state-of-the-art for autonomous driving LiDAR perception.
\end{itemize}

\section{PREVIOUS WORK}

A number of deep-learning-based approaches to object detection and classification for LiDAR point clouds have been proposed.  In this section we describe several methods that are most relevant to ours.

PIXOR~\cite{yang2018cvpr:pixor} uses 2D convolutions to detect objects in LiDAR point clouds projected as multi-channel BEV tensors. The network computes object detection confidence maps and regresses bounding box positions, sizes, and orientations for each output pixel. Bounding boxes are extracted by clustering. The method is fast, but challenging objects like pedestrians can be misdetected due to similarities with other objects in the top-down view. In followup work, Fast and Furious~\cite{luo2018cvpr:fastnfurious} addresses simultaneous detection, object tracking, and motion forecasting. The approach is similar to PIXOR, but some versions are slow due to their use of 3D convolutions applied to the voxelized point cloud (4D tensor). Results are shown only for an internal dataset. 

VoxelNet~\cite{zhou2018cvpr:voxelnet} voxelizes the entire 3D LiDAR point cloud, which is randomly subsampled for dense voxels. A region proposal network (RPN) is used for 3D object detection, and voxels are converted into dense 3D tensors. This leads to challenges with slow running time, and information loss due to voxelization. Similarly, SECOND~\cite{yan2018sensors:second} converts the point cloud to voxel features and coordinates, then applies sparse convolution, followed by RPN. Information is extracted vertically from the point cloud before downsampling 3D data. Runtime is improved, but information is still lost due to voxelization.

MV3D~\cite{chen2017cvpr:mv3d} converts the point cloud to a multi-channel BEV representation with several channels representing height map and intensity.  Both LiDAR and RGB camera data are used together to improve accuracy. The point cloud is projected onto BEV and ego-centric views, LiDAR and RGB camera data are combined together, and region proposals are used for 3D detection. Similarly, AVOD~\cite{ku2018iros:avod} also uses both LiDAR and RGB, along with a novel feature extractor based on feature pyramid network (FPN)~\cite{tsungyilin2017cvpr:fpn} and RPN for object detection.

PointNet~\cite{qi2017cvpr:pointnet} performs object classification and per-point semantic segmentation, operating directly on an unordered 3D point cloud. PointPillars~\cite{lang2019cvpr:pointpillars} uses only the LiDAR point cloud, achieving state-of-the-art accuracy and speed by using PointNets to organize a point cloud into columns. The network encodes points, runs a simplified version of PointNet, and finally uses SSD~\cite{weiliu2016eccv:ssd} to detect objects. STD~\cite{yang2019arx:stdobj} uses two stages:  PointNet++ for semantic segmentation of a point cloud and proposal generation network for classification and regression predictions.

The recently proposed Lidar DNN~\cite{zhou2019corl:ete} also uses perspective and top-down BEV views of LiDAR point clouds. Bounding boxes are used as supervision targets, and two parallel branches of the network learn to extract features relevant to object detection from each view via 2D convolutions. These features, which are learned implicitly, are then used to detect objects.  Like other methods, the results appear to show single class object detection.  
In contrast, our approach uses explicit features and representations for perspective and top-down view, which makes the system easy to train and debug; our network performs multi-class object detection and is an order of magnitude faster.

RangeNet++~\cite{milioto2019iros:rangenetpp} spherically projects the point cloud onto an ego-centric range image, on which a 2D semantic segmentation network is applied.  A fast, GPU-accelerated $k$-nearest neighbor ($k$NN) post-processing step is applied to the unprojected segmented point cloud to clean up the effects of bleeding between adjacent objects.  
The result is a real-time approach that can semantically label all points of the original point cloud, regardless of the discretization level of the network. The network is trained on SemanticKITTI, which is a version of the KITTI dataset~\cite{geiger2012cvpr:kitti,geiger2013ijrr:kitti} where each LiDAR point is semantically labeled for 25 classes. 


\section{METHOD}

\begin{figure*}
\centering
	\begin{tabular}{c}
	\includegraphics[width=1.5\columnwidth]{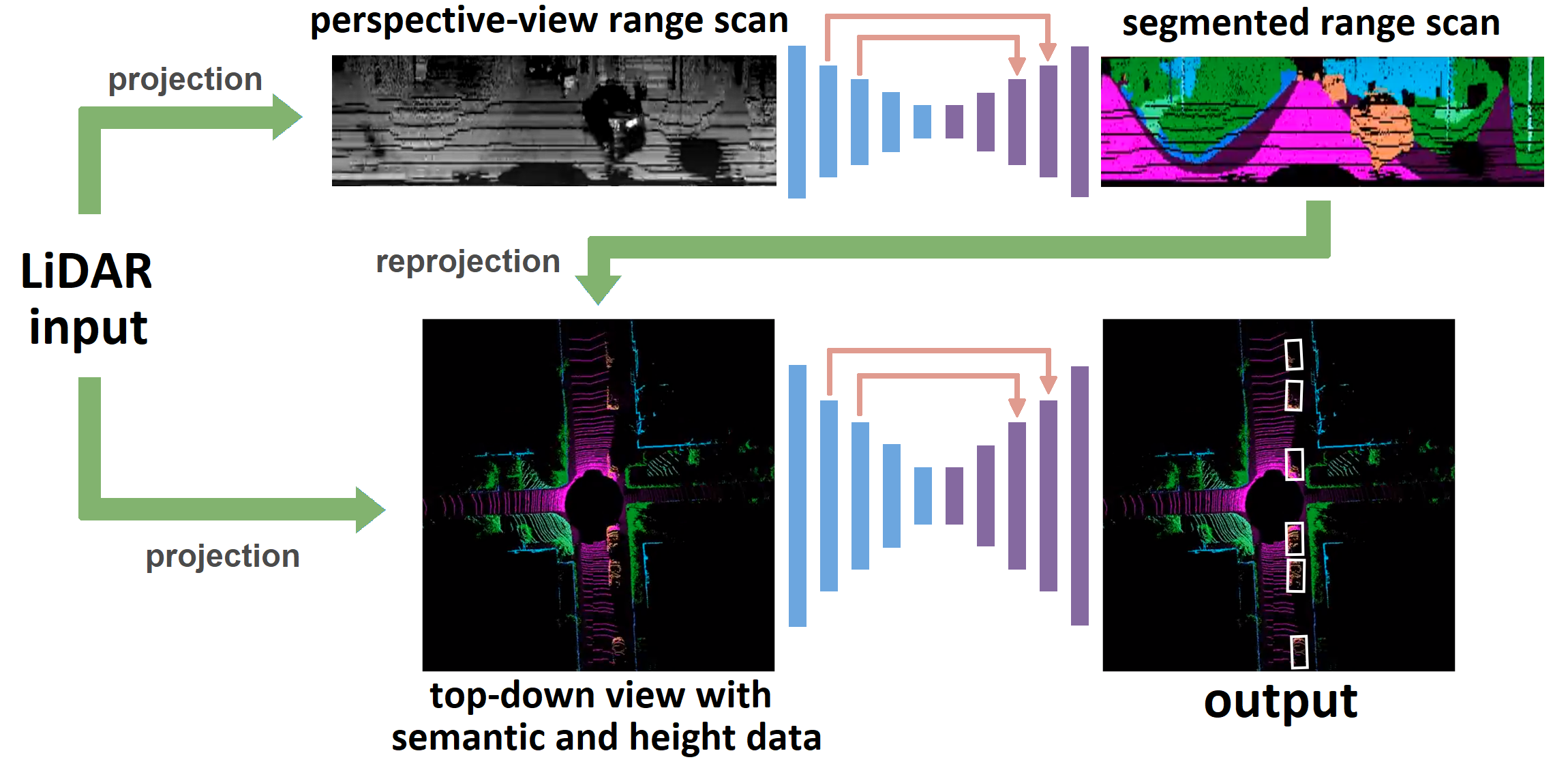}
	\end{tabular}
\caption{Our proposed MVLidarNet is a neural network with two stages. The first stage performs semantic segmentation (including drivable space) on the LiDAR input after projecting to a perspective view. The second stage uses the output of the first stage reprojected to a top-down view, along with the LiDAR input height map, to detect dynamic objects.  Both stages are feature pyramid networks (FPNs).}
\label{fig:mvlidar-net}
\end{figure*}

The proposed system consists of a two-stage neural network, as shown in Fig.~\ref{fig:mvlidar-net}.
The input to the system is a motion-compensated LiDAR point cloud capturing a 360$^\circ$ view of the scene, which is projected both perspectively (spherical projection) and top-down (orthographic projection).

The first stage extracts semantic information from the perspectively-projected LiDAR range scan. 
For the sensors used in this work, the resolution of each scan is $n_v \times n_s$, where $n_v=64$ is the number of horizontal lines, and $n_s=2048$ is the number of samples per line.
Each sample contains the distance to the corresponding point as well as the intensity of the received signal.
From these distances and intensities, the first stage segments the range scan into the following 7 classes: cars, trucks, pedestrians, cyclists, road surface, sidewalks, and unknown.
(We experimented with more or fewer classes but found the best results were obtained with this choice.)
Our motivation for using a perspective view is that pedestrians and cyclists are more easily discerned in this manner due to their characteristic shapes.


The architecture is similar to the feature pyramid network (FPN) \cite{tsungyilin2017cvpr:fpn} in its design. 
Processing is fast, since only 2D convolution/deconvolution layers are used. 
As detailed in Table~\ref{tab:arch1}, the encoder consists of three convolutional layers with 64, 64, and 128 $3 \times 3$ filters at the input resolution, followed by three Inception blocks~\cite{szegedy2016cvpr:inception} at decreased resolutions (using max pooling for downsampling).
The three Inception blocks consist of two, two, and three Inception modules, respectively.\footnote{Specifically, we use Inception-v2 modules, as depicted in Fig.~5 of~\cite{szegedy2016cvpr:inception}.} 
The decoder brings features back to the original resolution with 3 deconvolution blocks.
Each block consists of a deconvolution layer followed by two convolution layers with $1 \times 1$ and $3 \times 3$ filters, respectively. 
The number of filters within these blocks (across all layers) is 256, 128, and 64, respectively.
Skip connections are added at quarter and half resolution. 
The classification head consists of a $3 \times 3$ convolution layer with output feature size of 64, followed by a $1 \times 1$ layer that outputs a 7-element vector per pixel indicating the class probabilities.
Every convolution layer is followed by batch normalization and ReLU activation. Input and output have the same spatial resolution, and cross-entropy is used as the loss in training.  Note that this first stage directly outputs the drivable space.


\begin{table}%
\centering
\caption{Network architecture of first stage.  Output dimensions are (depth, height, width).  Final output is per-pixel semantic score for 7 classes.}
\begin{small}
\begin{tabular}{l|l|l}
\textbf{Name} & \textbf{Layer description} & \textbf{Output dim.} \\
\hline
\hline
& \emph{Input LiDAR scan} & \emph{(3, 64, 2048)} \\
trunk1 & conv (3$\times$3) & (64, 64, 2048) \\
trunk2 & conv (3$\times$3) & (64, 64, 2048) \\
trunk3 & conv (3$\times$3), 2$\downarrow$ & (128, 32, 1024) \\
block1 & 2$\times$Inception & (64, 32, 1024) \\
block2 & 2$\times$Inception, 2$\downarrow$ & (64, 16, 512) \\
block3 & 3$\times$Inception, 2$\downarrow$ & (128, 8, 256) \\
\hline
up1a & deconv, 2$\uparrow$ & (256, 16, 512) \\
up1b & concat block2, up1a & (256+64, 16, 512) \\
up1c & conv (1$\times$1) & (256, 16, 512) \\
up1d & conv (3$\times$3) & (256, 16, 512) \\
up2a & deconv, 2$\uparrow$ & (128, 32, 1024) \\
up2b & concat block1, up2a & (128+64, 32, 1024) \\
up2c & conv (1$\times$1) & (128, 32, 1024) \\
up2d & conv (3$\times$3) & (128, 32, 1024) \\
up3a & deconv, 2$\uparrow$ & (64, 64, 2048) \\
up3b & conv (1$\times$1) & (64, 64, 2048) \\
up3c & conv (3$\times$3) & (64, 64, 2048) \\
\hline
classhead1 & conv (3$\times$3) & (64, 64, 2048) \\
classhead2 & conv (1$\times$1) & \textbf{(7, 64, 2048)} \\
\end{tabular}
\end{small}
\label{tab:arch1}
\end{table}

\begin{table}%
\centering
\caption{Network architecture of second stage.  Input includes top-down projected per-pixel semantic score for 7 classes, along with top-down projected LiDAR (min height, max height, and mean intensity).  Final output is (downsampled) per-pixel score for 3 classes, along with 6 bounding box parameters.}
\begin{small}
\begin{tabular}{l|l|l}
\textbf{Name} & \textbf{Layer description} & \textbf{Output dim.} \\
\hline
\hline
& \emph{Reprojected semantics} & \emph{(7, 1024, 1024)} \\
sem1 & conv (3$\times$3) & (16, 1024, 1024) \\
sem2 & conv (3$\times$3) & (16, 1024, 1024) \\
sem3 & conv (3$\times$3), 2$\downarrow$ & (32, 512, 512) \\
sem4 & conv (3$\times$3) & (32, 512, 512) \\
\hline
	   & \emph{Reprojected LiDAR} & \emph{(3, 1024, 1024)} \\
height1 & conv (3$\times$3) & (16, 1024, 1024) \\
height2 & conv (3$\times$3) & (16, 1024, 1024) \\
height3 & conv (3$\times$3), 2$\downarrow$ & (32, 512, 512) \\
height4 & conv (3$\times$3) & (32, 512, 512) \\
\hline
block0 & concat sem4, height4 & (32+32, 512, 512) \\
block1a & conv (3$\times$3) & (64, 512, 512) \\
block1b & conv (3$\times$3), 2$\downarrow$ & (64, 256, 256) \\
block2a & conv (3$\times$3) & (128, 256, 256) \\
block2b & conv (3$\times$3), 2$\downarrow$ & (128, 128, 128) \\
block3a & conv (3$\times$3) & (256, 128, 128) \\
block3b & conv (3$\times$3), 2$\downarrow$ & (256, 64, 64) \\
\hline
up1a & deconv, 2$\uparrow$ & (128, 128, 128) \\
up1b & concat block2b, up1a & (128+128, 128, 128) \\
up1c & conv (3$\times$3) & (128, 128, 128) \\
up2a & deconv, 2$\uparrow$ & (64, 256, 256) \\
up2b & concat block1b, up2a & (64+64, 256, 256) \\
up2c & conv (3$\times$3) & (64, 256, 256) \\
\hline
classhead1 & conv (3$\times$3) from up2c & (64, 256, 256) \\
classhead2 & conv (3$\times$3) & (32, 256, 256) \\
classhead3 & conv (3$\times$3) & \textbf{(3, 256, 256)} \\
\hline
bboxhead1 & conv (3$\times$3) from up2c & (64, 256, 256) \\
bboxhead2 & conv (3$\times$3) & (32, 256, 256) \\
bboxhead3 & conv (3$\times$3) & \textbf{(6, 256, 256)} 
\end{tabular}
\end{small}
\label{tab:arch2}
\end{table}


%

The semantically labeled scan is reprojected onto a top-down view and combined with height information from the projected LiDAR data. The resulting combination is used by the second stage.
Using class probabilities (rather than the most likely class) enables the network to perform more complex reasoning about the data (e.g., a person on a bicycle); we experimented with both and found this approach to yield better results.

The second stage architecture (see Table~\ref{tab:arch2}) is also an encoder-decoder with skip connections, with two heads for classification and bounding box regression. We use $3 \times 3$ 2D convolution/deconvolution in all layers, which again yields fast processing times. The encoder starts with two parallel input blocks for semantics and height data (where the latter contains min height, max height, and mean intensity). Each block has 4 layers with 16, 16, 32, and 32 filters, respectively.  Each input block downsamples by 2 spatially, and the resulting tensors are concatenated along the depth dimension and fed to the encoder consisting of 3 blocks. Each of these blocks has 2 layers and downsamples the input by 2 spatially. The blocks have 64, 128, and 256 filters, respectively. The decoder consists of 2 blocks. Each block has a deconvolution (upsample by 2) followed by a convolution layer. Blocks have 128 and 64 filters, respectively. We add skip connections to each deconvolution from the corresponding encoder layer. The pixel classification head has 3 layers with 64, 32, and $n_c$ number of filters, while the bounding box parameters regression head has 3 layers with 64, 32, and $n_r$ number of filters (where $n_c$ and $n_r$ are described below). We use batch normalization followed by ReLU activation for all but the task head layers. 

The final output consists of a $256 \times 256$ array of $n_c$-element vectors containing the class distribution ($n_c=3$: vehicle, pedestrian, and unknown), along with another $256 \times 256$ array of $n_r$-element vectors ($n_r=6$), each containing $[\delta_x,\delta_y,w_o,\ell_o,\sin\theta,\cos\theta]$, where $(\delta_x,\delta_y)$ points toward the centroid of the corresponding object, $w_o \times \ell_o$ are the object dimensions, and $\theta$ is the orientation in the top-down view.
By representing the object dimensions in this parameterized manner, our vehicle detection includes not only cars but also buses and trucks.  
The stage is trained with a loss that combines focal loss~\cite{lin2017iccv:focal} for the classification head and L1 loss for the regression head, with corresponding weights.

A clustering algorithm is applied to the output of the second stage to detect individual instances of objects.
For clustering, we use the DBSCAN algorithm~\cite{ester1996kdd:dbscan} applied to the regressed centroids using only the cells whose class confidence produced by the classification head exceeds a threshold.
Individual bounding box coordinates, dimensions, and orientations are averaged within each cluster.

Note that the simplicity of the proposed approach makes it easy to implement and computationally efficient.
The method leverages the best of both worlds: The perspective view captures rich shape information that allows semantic understanding of the scene, while the top-down view allows metric reasoning without the difficulties in handling occlusion. Both views are processed using only 2D convolutions, which are much faster than 3D convolutions that are commonly used with voxelized volumes. 

The two stages can be trained either independently or together.  
We train them separately due to the non-overlapping nature of existing segmentation and object detection datasets (e.g., SemanticKITTI~\cite{behley2019iccv:semantickitti} and KITTI~\cite{geiger2012cvpr:kitti,geiger2013ijrr:kitti}).
If, however, appropriately labeled data were available, both stages could be trained together end-to-end. 

\begin{figure}
	\includegraphics[width=\columnwidth]{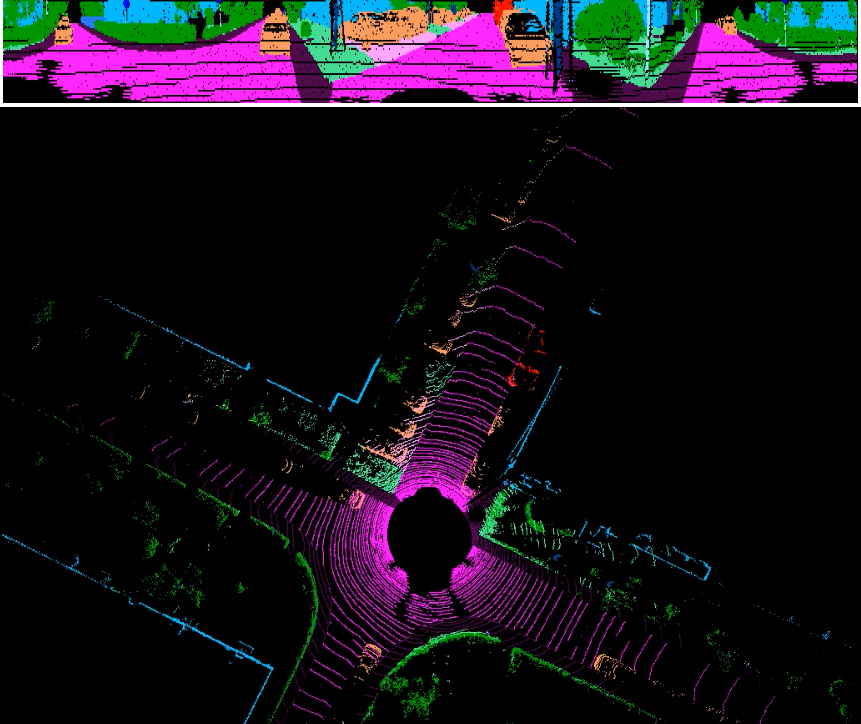}
	\caption{Top: Segmentation output of the first stage on a frame of the KITTI dataset. Bottom: BEV reprojection of the semantic segmentation and height data from the first stage, used as input to the second stage.}
	\label{fig:lidar-segnet-1ststage}
\end{figure}

\begin{figure*}
	\centering
	\begin{tabular}{c}
		\includegraphics[width=2\columnwidth]{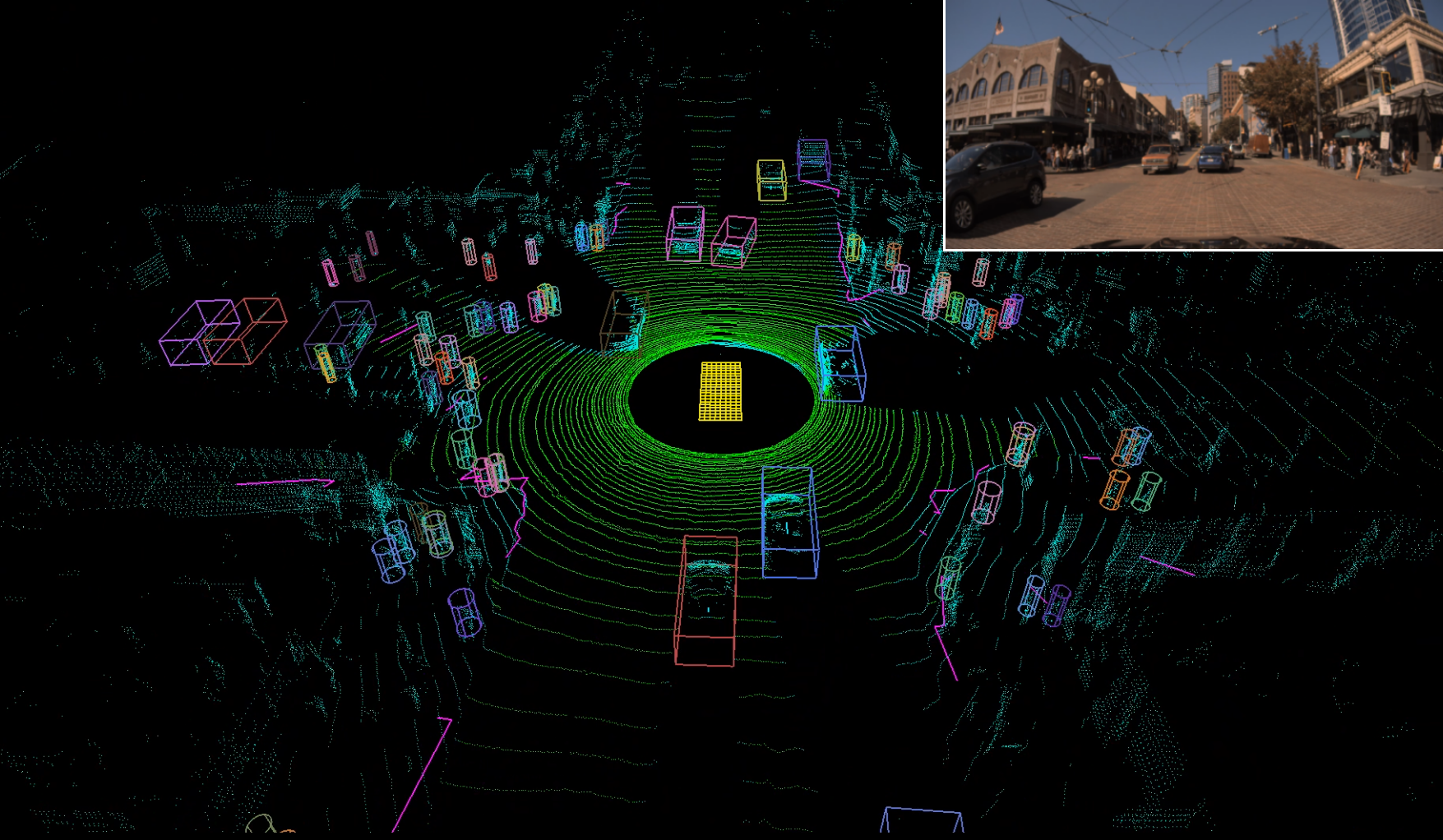}
	\end{tabular}
	\caption{End-to-end object detection and segmentation results for vehicles (cars, buses, and trucks), pedestrians, drivable space (green) on a crowded urban scene. 133 objects were detected in this frame:  vehicles (boxes) and pedestrians (cylinders).  Different colors indicate different instances, which are tracked over time for label consistency.  Box and cylinder heights are determined by a postprocessing step.}
	\label{fig:results-crowded-endtoend}
\end{figure*}

\section{EXPERIMENTAL RESULTS}

In this section we show that our simple approach, which consists of a single multi-class network trained on data from a single Velodyne HDL-64E, is able to achieve competitive results with an order of magnitude less computation than competing approaches.  Moreover, we show results detecting vehicles and pedestrians in crowded scenes containing more than one hundred objects in each frame.  (We do not show cyclist results, as the datasets do not contain enough cyclists to train without data augmentation.)  As mentioned above, the two stages are trained independently due to differences in segmentation and object detection datasets (e.g. KITTI vs.~SemanticKITTI). The first segmentation stage is trained to segment class masks only and uses segmentation labels either from labeled LiDAR point clouds directly or segmentation labels transferred from camera to LiDAR. The second stage is trained on LiDAR bounding box labels. 

The input LiDAR data is motion compensated for training and inference. The input/output resolution of the first stage is set to $64 \times 2048$. The input resolution of the second stage is set to $w=1024$ and $\ell=1024$ as a compromise between spatial resolution, cell occupancy, and computational load. It covers an $80 \times 80$~m$^2$ area and yields a cell resolution of $7.8$~cm per cell. The output of the second stage is set to $256 \times 256$, with each output cell having a spatial resolution of $31.3$~cm.

We use the Adam optimizer with initial learning rate set to $10^{-4}$. For the second stage loss, we set the class and regression weights to 5.0 and 1.0, respectively. Both stages are trained for 40 to 50 epochs with a batch size of 4. 
The resulting model is exported to TensorRT for inference on the vehicle using an embedded GPU.

\subsection{Semantic segmentation}

\begin{table*}%
\caption{LiDAR segmentation results for SemanticKITTI~\cite{behley2019iccv:semantickitti} dataset, comparing our segmentation network with RangeNet++~\cite{milioto2019iros:rangenetpp}. Shown are the IoU scores for different classes, and computation time.  Input size is $64 \times 2048$.}
\begin{small}
\begin{tabular}{c|p{0.1in}p{0.1in}p{0.1in}p{0.1in}p{0.1in}p{0.1in}p{0.1in}p{0.1in}p{0.1in}p{0.1in}p{0.1in}p{0.1in}p{0.1in}p{0.1in}p{0.1in}p{0.1in}p{0.1in}p{0.1in}p{0.2in}|p{0.2in}|c}
Method & \rotatebox{90}{car} & \rotatebox{90}{bicycle} & \rotatebox{90}{motorcycle} & \rotatebox{90}{truck} & \rotatebox{90}{other-vehicle} & \rotatebox{90}{person} & \rotatebox{90}{bicyclist} & \rotatebox{90}{motorcyclist} & \rotatebox{90}{road} & \rotatebox{90}{parking} & \rotatebox{90}{sidewalk} & \rotatebox{90}{other-ground} & \rotatebox{90}{building} & \rotatebox{90}{fence} & \rotatebox{90}{vegetation} & \rotatebox{90}{trunk} & \rotatebox{90}{terrain} & \rotatebox{90}{pole} & \rotatebox{90}{traffic-sign} & \rotatebox{90}{mean IoU} & \rotatebox{90}{speed (fps)} \\
\hline
RangeNet53 & 86.4 & 24.5 & 32.7 & 25.5 & 22.6 & 36.2 & 33.6 & 4.7 & \textbf{91.8} & 64.8 & 74.6 & \textbf{27.9} & 84.1 & 55.0 & 78.3 & 50.1 & 64.0 & 38.9 & 52.2 & 49.9 & 13 \\
RangeNet53++ & \textbf{91.4} & 25.7 & \textbf{34.4} & \textbf{25.7} & 23.0 & 38.3 & 38.8 & 4.8 & \textbf{91.8} & \textbf{65.0} & \textbf{75.2} & 27.8 & \textbf{87.4} & \textbf{58.6} & 80.5 & 55.1 & \textbf{64.6} & 47.9 & \textbf{55.9} & 52.2 & 12 \\
\hline
Ours & 86.3 & 33.8 & 34.2 & 24.0 & \textbf{25.4} & 44.0 & 41.8 & 23.0 & 90.3 & 56.5 & 72.6 & 19.4 & 83.0 & 51.2 & 79.0 & 54.9 & 63.4 & 41.9 & 52.8 & 51.5 & \textbf{200} \\
Ours++ & 87.1& \textbf{34.9} & 32.9 & 23.7 & 24.9 & \textbf{44.5} & \textbf{44.3} & \textbf{23.1} & 90.3 & 56.7 & 73.0 & 19.1 & 85.6 & 53.0 & \textbf{80.9} & \textbf{59.4} & 63.9 & \textbf{49.9} & 51.1 & \textbf{52.5} & 92
\end{tabular}
\end{small}
\label{tab:segnet}
\end{table*}


Table~\ref{tab:segnet} provides comparisons of our first stage segmentation results with RangeNet LiDAR segmentation network on SemanticKITTI~\cite{behley2019iccv:semantickitti} dataset. Our network runs much faster (200~fps vs.~13~fps) while providing similar accuracy (51.5 mIoU vs.~49.9 mIoU). At $64 \times 2048$ resolution, our end-to-end runtime is 2.1~ms in FP16 mode and 4.9~ms in FP32 mode on an NVIDIA Drive AGX computer. This is 200 frames per second, compared with RangeNet's best result of 13 frames per second at this resolution in FP32 mode. Our segmentation network runs at 480 frames per second in FP16 mode. We achieve this speed by using a simpler and shallower network structure based on the Inception architecture. The \emph{++} addition to methods in Table~\ref{tab:segnet} indicate extra post-processing as described in RangeNet++~\cite{milioto2019iros:rangenetpp}.

\subsection{Object detection}

\begin{table*}%
\centering
\caption{Evaluation on KITTI~\cite{geiger2012cvpr:kitti,geiger2013ijrr:kitti} BEV object detection benchmark (test split).  Shown are AP scores at 0.7 IoU for cars.}
\begin{small}
\begin{tabular}{c|c|c|c|c|r}
Method & Modality & Easy & Mod. & Hard & Speed (ms) \\
\hline
MV3D~\cite{chen2017cvpr:mv3d} & RGB + LiDAR & 86.02 & 76.90 & 68.49 & 240 \\
ContFuse~\cite{liang2018eccv:contfuse} & RGB + LiDAR & 88.81 & 85.83 & 77.33 & 60 \\
AVOD-FPN~\cite{ku2018iros:avod} & RGB + LiDAR & 88.53 & 83.79 & 77.90 & 100 \\
F-PointNet~\cite{qi2018cvpr:fpointnet} & RGB + LiDAR & 88.70 & 84.00 & 75.33 & 170 \\
MMF~\cite{liang2019cvpr:mtmsfobj} & RGB + LiDAR & 89.49 & 87.47 & 79.10 & 80 \\
\hline
VoxelNet~\cite{zhou2018cvpr:voxelnet} & LiDAR only & 89.35 & 79.26 & 77.39 & 500 \\
SECOND~\cite{yan2018sensors:second} & LiDAR only & 88.07 & 79.37 & 77.95 & 40 \\
PointPillars~\cite{lang2019cvpr:pointpillars} & LiDAR only & 88.35 & 86.10 & 79.83 & 16 \\
PointRCNN~\cite{shi2019cvpr:pointrcnn} & LiDAR only & 89.47 & 85.68 & 79.10 & 100 \\
Part-$A^2$~\cite{shi2019arx:partatwo} & LiDAR only & 89.52 & 84.76 & 81.47 & 80 \\
STD \cite{yang2019arx:stdobj} & LiDAR only & \textbf{89.66} & \textbf{87.76} & \textbf{86.89} & 80 \\
Ours & LiDAR only & 89.27 & 80.59 & 70.90 & \textbf{7}
\end{tabular}
\end{small}
\label{tab:objectKITTI}
\end{table*}

\begin{table*}%
	\centering
	\caption{Evaluation on our large internal LiDAR object detection dataset. Shown are AP scores for BEV detections at various ranges for vehicles (cars, trucks, buses) and pedestrians. We compare our two-stage semantic segmentation + top-down detection with a top-down approach relying on height information alone. Our semantics+height MVLidarNet approach achieves significantly better results.}
	\begin{small}
		\begin{tabular}{c|c|c|c|c|c}
			Method & Class & IoU & 0--10~m & 10--25~m & 25--50~m \\
			\hline
			Top-down height only & Vehicles & 0.7 & 96.45 & \textbf{91.52} & 77.48 \\
			Top-down height only & Pedestrians & 0.5 & 51.75 & 48.20 & 28.67 \\
			\hline
			Perspective semantics + top-down height & Vehicles & 0.7 & \textbf{96.80} & 91.20 & \textbf{77.77} \\
			Perspective semantics + top-down height & Pedestrians & 0.5 & \textbf{72.29} & \textbf{59.01} & \textbf{39.17} 
		\end{tabular}
	\end{small}
	\label{tab:internalres}
\end{table*}

Table \ref{tab:objectKITTI} provides comparisons of our full network's output for bird's eye view (BEV) object detection on KITTI~\cite{geiger2012cvpr:kitti,geiger2013ijrr:kitti} dataset for cars. Our network runs much faster than others (7~ms vs.~16~ms for the next fastest) while providing competitive accuracy (AP score of 89.27 vs.~89.66 for the best algorithm on easy cars). Our end-to-end (two stages combined) runtime is 6.8~ms per frame (corresponding to about 150~fps).

Since the KITTI dataset does not have many pedestrian instances for training (4487 instances in the object detection set), PointPillars~\cite{lang2019cvpr:pointpillars} introduced a set of augmentation techniques to improve AP scores on KITTI. Such techniques are not reflective of the actual data and therefore potentially bias the network since they do not capture real LiDAR geometry. As a result, we do not use these augmentation techniques. We only use horizontal flips and global rotations.  Nevertheless, our approach still achieves competitive results due to the simple network architecture and semantic segmentation.  

\begin{figure}
\begin{tabular}{cc}
	\hspace{-1em} \includegraphics[width=0.5\columnwidth]{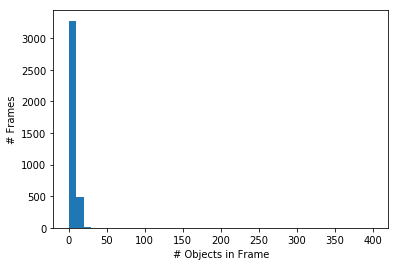} &
	\hspace{-1.5em} \includegraphics[width=0.5\columnwidth]{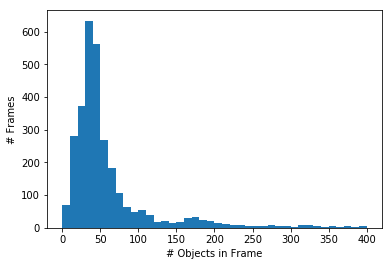} \\
	KITTI & Ours
\end{tabular}
	\caption{Histogram showing the number of objects per image frame for KITTI (left) and our internal dataset (right).  Whereas KITTI never contains more than 25 objects in a single frame, our dataset contains hundreds of frames with more than 100 objects.}
	\label{fig:hist-kitti-ours}
\end{figure}

Our internal LiDAR dataset is much larger than KITTI (hundreds of thousands of scans) and includes more challenging scenes with crowds of pedestrians---nearly a hundred per frame in many cases (see Figure \ref{fig:hist-kitti-ours}). In total the dataset has 123,195 pedestrians. 
Table \ref{tab:internalres} provides AP scores computed on this internal dataset for vehicles and pedestrians. 
Note that these scores are from a single multi-class network, \emph{i.e.}, we do not train separately for vehicles and pedestrians but rather jointly.  It also shows an ablation study that compares our full network (two stages that use semantic segmentation and top-down detection) with just a top-down network that relies only on height data (\emph{i.e.}, the second stage was modified to take only height data as input). Using two stages with both semantics and height clearly shows significant improvement in pedestrian detection (72.29 vs.~51.75 at less than 10~m, \emph{etc.}). Thus, adding semantic segmentation helps with object detection on LiDAR point clouds. 

We also have collected an internal dataset with the higher-resolution Velodyne VLS-128.  On this dataset, our network achieves even higher accuracy:  97.48, 95.39, and 86.77 (vehicles); and 88.89, 61.27, and 48.34 (pedestrians).

An example of our multi-class detection is shown on Figure~\ref{fig:results-crowded-endtoend}. Note that in addition to detected dynamic objects, our system marks segmented drivable space (in green color vs.~other LiDAR points in cyan). Currently, we only use drivable space, but with sufficient training data, other scene elements like sidewalks, trees, buildings, and poles provided by the first stage could be added. Such semantic features are not possible with standard object detection networks.

\section{CONCLUSION}

We have presented a multi-class, multi-view network that simultaneously detects dynamic objects (vehicles and pedestrians) and segments the drivable space from LiDAR point cloud data. Our network consists of two stages: 1) one stage that semantically labels the points in the LiDAR range scan in a perspective view, and 2) another stage that detects objects using semantically segmented points reprojected onto a top-down bird's eye view (BEV), combined with height data from the LiDAR point cloud. 
This simple architecture achieves results that are competitive with state-of-the-art with considerably less computation.
Moreover, results are achieved on crowded scenes with unprecedented complexity, with a single multi-class network, and without relying on complex data augmentation schemes used by previous techniques.
Future work includes training the two stages end-to-end with a combined segmentation and object detection dataset, experimenting with different height and semantics encodings, and extending the number of supported classes.

\bibliographystyle{IEEEtran}
\bibliography{refs}

\end{document}